\documentclass{ecai} 

\usepackage{latexsym}
\usepackage{amssymb}
\usepackage{amsmath}
\usepackage{amsthm}
\usepackage{booktabs}
\usepackage{enumitem}
\usepackage{graphicx}
\usepackage{color}
\usepackage{hyperref}
\usepackage{bbold}
\usepackage{mathtools}
\usepackage{multicol}
\usepackage{multirow}
\usepackage{algorithm} 
\usepackage{algpseudocode} 
\usepackage{natbib}

\newcommand{\BibTeX}{B\kern-.05em{\sc i\kern-.025em b}\kern-.08em\TeX}
\definecolor{clemson-orange}{RGB}{234,106,32}
\definecolor{highlight-orange}{RGB}{255,150,150}
\definecolor{chicago-maroon}{RGB}{128,0,0}
\definecolor{cincinnati-red}{RGB}{190,0,0}
\definecolor{soft-cyan}{RGB}{68,85,90}
\definecolor{firebrick}{RGB}{178,34,34}
\definecolor{crimson}{RGB}{220,20,60}
\definecolor{cerrulean}{rgb}{0.165,0.322,0.745}
\definecolor{jaam}{rgb}{0.45,0.0,0.45}

\begin{document}


\begin{frontmatter}


\title{Noise-Free Explanation for Driving Action Prediction}


\author[A]{\fnms{Hongbo}~\snm{Zhu}\thanks{Corresponding Author. Email:hongbo.zhu@manchester.ac.uk.}}
\author[A]{\fnms{Theodor}~\snm{Wulff}}
\author[A]{\fnms{Rahul Singh}~\snm{Maharjan}}
\author[B]{\fnms{Jinpei}~\snm{Han}}
\author[A]{\fnms{Angelo}~\snm{Cangelosi}}

\address[A]{Manchester Centre for Robotics and AI, University of Manchester}
\address[B]{Brain \& Behavior Lab, Department of Computing, Imperial College London}


\begin{abstract}
Although attention mechanisms have achieved considerable progress in Transformer-based architectures across various Artificial Intelligence (AI) domains, their inner workings remain to be explored. Existing explainable methods have different emphases but are rather one-sided. They primarily analyse the attention mechanisms or gradient-based attribution while neglecting the magnitudes of input feature values or the skip-connection module. Moreover, they inevitably bring spurious noisy pixel attributions unrelated to the model's decision, hindering humans' trust in the spotted visualization result. Hence, we propose an easy-to-implement but effective way to remedy this flaw: \textbf{S}mooth \textbf{N}oise \textbf{N}orm \textbf{A}ttention (SNNA). We weigh the attention by the norm of the transformed value vector and guide the label-specific signal with the attention gradient, then randomly sample the input perturbations and average the corresponding gradients to produce noise-free attribution. Instead of evaluating the explanation method on the binary or multi-class classification tasks like in previous works, we explore the more complex multi-label classification scenario in this work, i.e., the driving action prediction task, and trained a model for it specifically. Both qualitative and quantitative evaluation results show the superiority of SNNA compared to other SOTA attention-based explainable methods in generating a clearer visual explanation map and ranking the input pixel importance. 
\end{abstract}

\end{frontmatter}


\section{Introduction}

The Vision Transformer (ViT) is becoming a prominent architecture in computer vision (CV) tasks \cite{khan2022transformers} like image classification, segmentation and object detection. ViT is effective at processing spatial and temporal data, outperforming traditional convolutional neural networks (CNNs) by removing the assumption imposed by kernel size, and makes enormous progress in autonomous driving tasks. However, The inner workings of Deep Learning (DL) models, including Transformers, remain opaque \cite{mohebbi2024transformer}, necessitating the development of eXplainable AI (XAI) methods to make their decision-making processes more transparent. XAI aims to make deep neural networks (DNNs) more understandable and reliable by providing insights into why the model makes specific decisions. This transparency enhances trust in AI systems, enables the detection of biases or errors, and facilitates alignment with ethical and legal standards \cite{sigfrids2023human}. Visualizing the decision process of DL models is crucial for debugging downstream tasks, ultimately enhancing their applicability and trustworthiness in real-world scenarios \cite{DBLP:journals/corr/abs-2204-10464}.

\begin{figure}[ht!]
\centering
\includegraphics[scale=0.55]{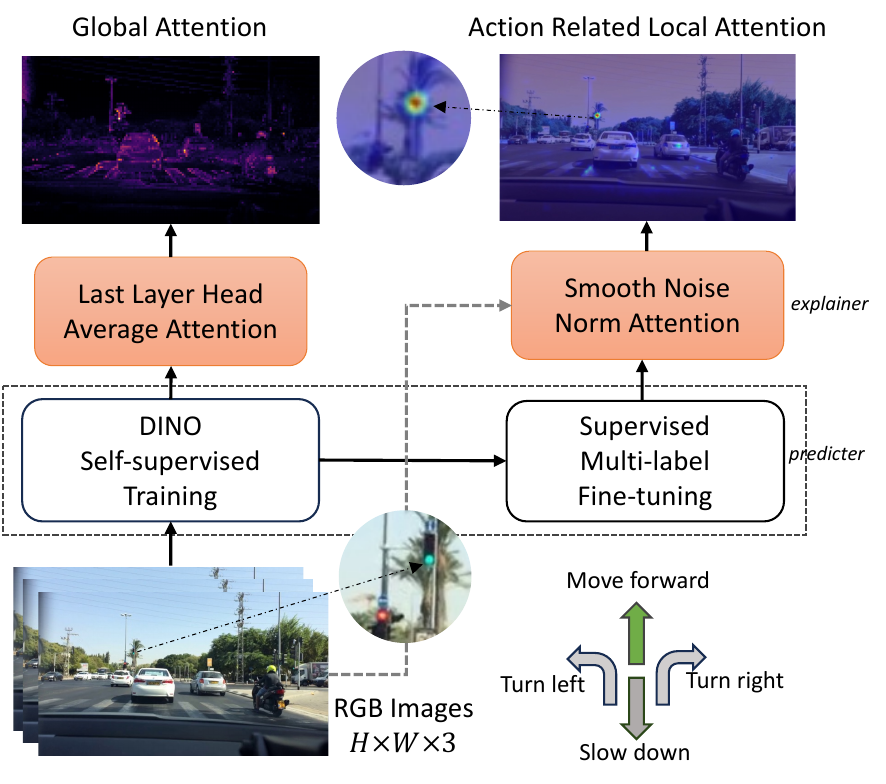}
\caption{Our workflow consists of first training the predictor and then using our explainer (SNNA) to generate a noise-free visual explanation.}
\label{fig:framework}
\end{figure}

The self-attention mechanism in transformers assigns weights between input tokens, indicating their relative importance. This enables the use of attention weights in the analysis of model outputs by revealing an importance distribution over the input space \cite{chaudhari2021attentive}. However, attention score alone does not provide a comprehensive understanding of model behaviour, e.g., latent activations. Also, attention lacks class-specificity and primarily indicates the softmax output (in attention layers), neglecting other model components, e.g., the magnitudes of input feature values or the skip-connection module. Studies by \citet{serrano2019attention} demonstrate that removing representations with high attention weights does not consistently result in performance degradation. \citet{jain2019attention} found that attention scores may not serve as reliable explanations, as they often conflict with other indicators of feature importance, such as gradient-based measures. \citet{abnar2020quantifying} suggested that contextual information among tokens becomes more homogeneous as the model depth increases, undermining the reliability of explanations derived solely from raw attention weights. \citet{kobayashi2020attention} took the effect of input features into account but ignored the skip connection. \citet{chefer2021generic} considered the skip connection module, but no input features. Moreover, regardless of whether the attention-based or gradient-based method was used, none of them tried to solve the noise attribution problem, i.e., the presence of irrelevant or misleading visual highlights in the heatmap that do not correspond to the actual model's decision, which is non-trivial and cannot be ignored.

The necessity to explain driving action prediction arises from various factors. Firstly, societal expectations demand reliability assurances given the high-stakes and safety-critical nature of autonomous driving. Secondly, explanations are invaluable tools for enhancing system performance, allowing engineers and researchers to address deficiencies and glean insights into failure modes \cite{tian2018deeptest}. Thirdly, explanations foster trust among users, facilitating broader acceptance of the technology by providing transparency into the decision-making processes \cite{shen2020explain}. 
Driving is a complex task that involves various interacting factors. We may have multiple action choices in many driving scenarios, as shown in Appendix Figure \ref{fig:BDD-OIA}. So, driving action prediction involves more than simply assigning a single action label to each observed environment. Instead, it requires identifying all possible actions the system might take. This makes driving action prediction a multi-label classification problem.
Training deep learning models in a supervised way requires large amounts of labelled data, which is expensive and time-consuming to collect \cite{goldman2000enhancing}. Driving environment images from the cameras are easy to collect, while the driving action labels (i.e., move forward, slow down, turn left and turn right) are hard to get and are unavailable in most open-source autonomous driving datasets such as KITTI \cite{geiger2013vision}, ApolloScape \cite{huang2019apolloscape}, and Waymo \cite{sun2020scalability}, or only partially available in BDD-100k \cite{xu2020explainable}. It makes training a driving action prediction model in a purely supervised way extremely expensive since data needs to be annotated. \textbf{S}elf-\textbf{S}upervised \textbf{L}earning (SSL) \cite{jing2020self}, emerges as a compelling and flexible approach for training AI models, as it can leverage a vast amount of unlabelled data by learning a surrogate label or pre-text task created by the modeller, hence does not require extensive labelled data. 

Recent works have shown that we can learn unsupervised features without labels of the downstream task using SSL, which has seen huge interest in various Natural Language Processing (NLP) (BERT \cite{kenton2019bert}, GPT \cite{brown2020language} and T5 \cite{raffel2020exploring}) and CV (SimCLR \cite{chen2020simple}, MoCo \cite{DBLP:conf/cvpr/He0WXG20} and SwAV \cite{caron2020unsupervised}) models. These models utilize the widely available data without annotation for representation learning. Compared to supervised learning in CV, which simplifies the intricate visual details of an image to a single predefined category \cite{misra2020self}, SSL aims to assist the model in learning more transferable, generalizable and robust representations from pseudo-labels. To take advantage of the vast amount of unlabeled self-driving\footnote{Autonomous driving and self-driving express the same meaning, and we use them interchangeably in this paper.} images, we adapt DINO \cite{DBLP:conf/iccv/CaronTMJMBJ21} (Self-DIstillation with NO Labels) to train a driving action prediction model (ViT) using self-supervised learning on a large unlabelled dataset, followed by supervised learning on a smaller labelled dataset. Our pipeline is illustrated in Figure \ref{fig:framework}. DINO employs self-distillation, which trains a student model on small local crops and a teacher model on bigger crops of the input image, and then calculates the loss between both representations. It works by interpreting self-supervision as a particular case of self-distillation, where no labels are used. The teacher network will push the student network to learn global representations by seeing only the small local crops. DINO is related to co-distillation \cite{DBLP:conf/iclr/AnilPPODH18} where student and teacher have the same network architecture. Meanwhile, the teacher network in co-distillation also distils from the student part.

To overcome the noisy attribution problem, we introduce Smooth Noise Norm Attention (SNNA) \footnote{Our code can be found at \url{https://github.com/Hongbo-Z/SNNA}}, a concise and considered explainable method: we first weigh the attention weights with the norm of the transformed value vector, then mask with the attention gradient with respect to the prediction. Finally, the noise is filtered by sampling the instance around the input with Gaussian noise. By evaluating its performance in the driving action prediction task, the experimental results show that SNNA can identify the clear and sparse image pixel locations that lead to the predicted driving action.
To summarize, we state our contributions as follows:
\begin{itemize}
\item We proposed an easy-to-implement but effective noise-free explainable method, i.e., SNNA, and extend the current explanation boundary from multi-class to multi-label classification tasks.
\item We trained a model for the multi-label driving action prediction task, and conducted extensive experiments with five different baseline XAI methods, demonstrating SNNA is more faithful in identifying the decisive pixels for the model prediction than those methods under qualitative visualizations and two quantitative evaluation metrics, i.e., AUPC and LogOdd.
\end{itemize}

\section{Related Work}

\subsection{Explainable Autonomous Driving}
Understanding intelligent decision-making processes in autonomous vehicles remains challenging, impeding widespread societal acceptance of this technology \cite{ma2020artificial}. Therefore, in addition to ensuring safe real-time decision-making, autonomous vehicle systems should also explain how their decisions are formulated to comply with regulatory requirements across various jurisdictions \cite{omeiza2021explanations}. \citet{kim2017interpretable} utilize a causal attention model integrated with saliency filtering to identify input regions influencing the output, particularly in steering control. Building upon this work, \citet{kim2018textual} introduces textual explanations through an attention-based video-to-text mechanism, producing "explanations" for the decisive actions of self-driving vehicles. \citet{xu2020explainable} propose a paradigm for explainable reasoning of driving action-inducing objects. Most recently, \citet{jing2022inaction} developed an action decision-making model with explicit and implicit explanations. Though the above-mentioned methods provide some explanation with a saliency map, all these methods are designed for CNN-based models, not Transformer-based models. Additionally, these methods tend to highlight large areas of unrelated pixels, which confuses users a lot.

\subsection{Self-Supervised Learning \& Knowledge distillation}

SSL \cite{lan2019albert} is one of the main success factors in current NLP, which employs pretext tasks that leverage contextual information within sentences to provide a robust learning signal \cite{raffel2020exploring}. In CV, prevalent SSL strategies include pretext methods \cite{misra2020self} and contrastive learning \cite{jaiswal2020survey}; the former exploits inherent properties or tasks within the input data to learn representations, e.g., MAE \cite{DBLP:conf/cvpr/HeCXLDG22}, the latter encourages discriminative representations by maximizing dissimilarity between features extracted from different samples, e.g., CLIP \cite{radford2021learning}. For autonomous driving, SSL has been used for motion estimation \cite{bhattacharyya2023ssl}, and segmentation \cite{mayr2018self}. Knowledge distillation (KD) \cite{hinton2015distilling} compresses the knowledge learned by the teacher network into the student network, utilising soft labels (probability distributions). For the self-driving domain, KD has been used for driving scene semantic segmentation \cite{li2022self}, object detection \cite{lan2022adaptive}, and motion planning \cite{wang2021end}. However, to the best of our knowledge, neither SSL nor KD has yet been used for driving action prediction tasks among the works mentioned above. 

\begin{figure}[hb!]
\centering
\includegraphics[scale=0.45]{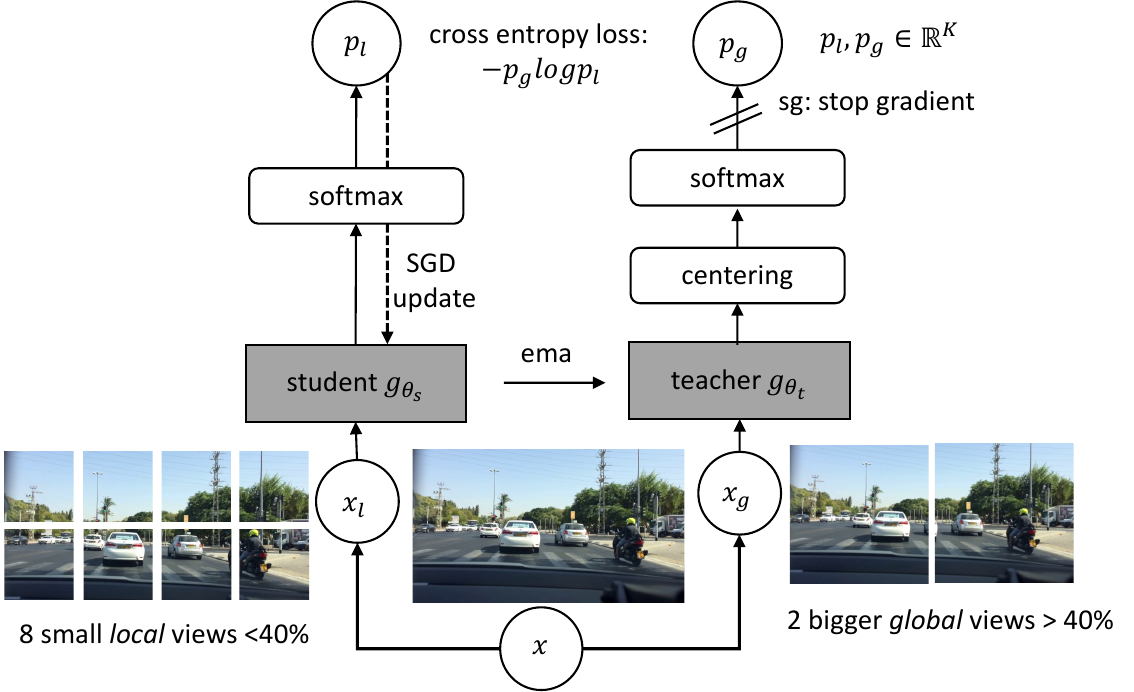}
\caption{Self-supervised training, where $g_{\theta_s}$ will learn from $g_{\theta_t}$. We randomly crop input image $x$ with $x_l$ smaller than 40\% of $x$ and $x_g$ bigger than 40\% of $x$. }
\label{fig:ssl}
\end{figure}

\subsection{Explainable Methods for Transformers}

\paragraph{Attention-based}
RawAtt \cite{jain2019attention} is the earliest proposed method that uses the last layer's attention weights as the salience of each input token. Based on RawAtt, Attention Rollout \cite{abnar2020quantifying} combines the attention weights across all layers to reassign the important scores to the tokens through the linear combination of attention weights across the layers, tracing the information flow in Transformer. However, the rollout operation cancels the accumulated importance scores as some deeper layers have almost uniformly distributed attention weights, ignoring the existing effects of linear projections inside attention blocks. AttGrad \cite{DBLP:conf/acl/ChrysostomouA20} considers both attention scores and how sensitive model prediction changes with respect to internal attention. AttIN \cite{kobayashi2020attention} takes the magnitude of transformed value vectors' norm into consideration. Generic Attribution\cite{chefer2021generic} generalizes the idea of Rollout and adds the gradient information to each layer's attention.

\paragraph{Gradient-based}
Gradient is one of the most straightforward approaches to depict the sensitivity of the trained model concerning each given feature as computed through backpropagation. InputGrad \cite{shrikumar2017learning} calculates the element-wise product between the input $x$ and the corresponding gradient: $x \odot \frac{\partial f}{\partial x}$. However, this computation is done on the average gradient and a linear interpolation of the input, which may incur the gradient saturation problem. Integrated Gradients (IG) \cite{sundararajan2017axiomatic} alleviates the gradient saturation issue by aggregating input gradients along the path between a baseline to the original input. However, it requires a selected baseline as a benchmark, which raises the question of how such a baseline should be chosen. IG's performance significantly depends on the selected baseline point. Choosing 0 (all-black image) as the baseline point inevitably leads to explanations that are insensitive to black objects, and this flaw is fatal to the self-driving scenario as black vehicles are quite common and black clothing is popular in different seasons.

Although the above methods have achieved prominent visual results in explaining model decisions, they suffer from remarkable flaws of contorted explanations and often create noisy attribution maps. Those visually striking noise pixels are scattered randomly across the saliency maps, as shown in the bottom row images of Figure \ref{fig:compare}. As suggested by \citet{sundararajan2017axiomatic}, the attribution method should satisfy the Sensitivity Axiom because the lack of sensitivity causes gradients to focus on irrelevant features. To tackle this problem, we draw inspiration from "remove noise by adding noise" \cite{smilkov2017smoothgrad} and propose "Smooth Noise Norm Attention" to avoid highlighting irrelevant pixels.

\section{Methodology}
As shown in Figure \ref{fig:framework}, our framework consists of two components, i.e., the trained predictor and the explainer SNNA. We first train the ViT in a self-supervised manner on a large unlabelled dataset, then fine-tune the classifier in a supervised manner on a relatively small human-labelled driving action dataset. Lastly, SNNA generates a visual explanation of the trained classifier. 

\subsection{Driving Action Predictor}
\subsubsection{Self-Supervised Learning }
We adapted DINO for the self-supervised training with ViT as the backbone network. In this setting, student and teacher models are trained simultaneously. The teacher model serves as a source of supervision for the student model. The student model learns to mimic the teacher model's behaviour using the teacher model's predictions as targets. As is illustrated in Figure \ref{fig:ssl}, we define the input image $x$, student network $g_{\theta_s}$ parameterized by $\theta_s$ and teacher network $g_{\theta_t}$ parameterized by $\theta_t$, while $x_l$ refers to the small local crops as input for $g_{\theta_s}$ and $x_g$ represents the bigger global crops as input for $g_{\theta_t}$. The student $g_{\theta_s}$ is trained by simply matching the output distribution of $g_{\theta_t}$ over different views (crops) of the same input image $x$. The neural network $g$ comprises a backbone $f$ (ViT) and a projection head $h$, i.e., $g=h \circ f$. $p_l$ and $p_g$ represent the student and teacher network's output probability distribution; the probability $p$ is obtained by normalizing the network's output $g$ with a softmax function. Both student and teacher networks share the same backbone architecture $g$ but with different parameters $\theta_s$ and $\theta_t$. The parameter $\theta_s$ is learned from $\theta_t$ with stochastic gradient descent. As teacher network $\theta_t$ is not priori known here, we build it from the last round of the student network $\theta_s$ using the "exponential moving average"(ema) and freezing the teacher network over an epoch. The teacher network performs better than the student throughout the training \cite{DBLP:conf/iccv/CaronTMJMBJ21}, guiding the student's training by providing target features of higher quality.

\begin{figure}
\centering
\includegraphics[scale=0.5]{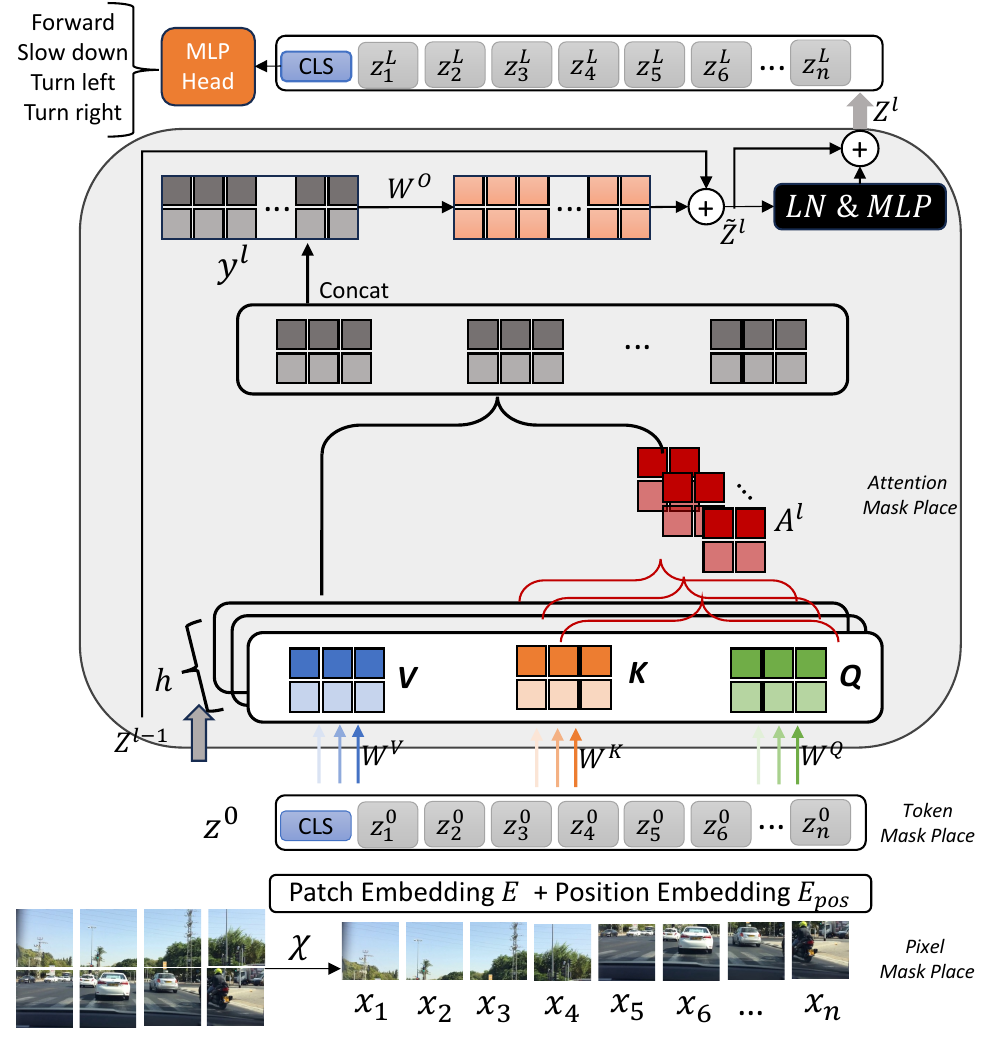}
\caption{An illustration of ViT architecture, the input image $\mathcal{X}$ is first split into $p\times p$ size small patches. Those patch sequences are mapped to Patch Embedding, and then a [CLS] token is prepended; after adding the Positional Encoding, we get the final input token sequence $Z^0$, which will be sent to the Transformer Encoder. Here, $h$ is the number of attention head, $A^l$ is the attention weights at layer $l$.}
\vspace{-1em}
\label{fig:vit}
\end{figure}

\paragraph{ViT} As is shown in Figure \ref{fig:vit}, an image is split into non-overlapping contiguous $p \times p$ pixels square patches in the ViT architecture. These patches are then passed through a linear transformation to form patch embeddings. This can be achieved by using a Conv2d with the kernel size and stride set to patch-size $p$. Then, an extra learnable token [CLS] is prepended to the embedded sequence, which serves as a representation of the entire image to perform classification, and we attach the projection head $h$ at its output. The patch tokens and the [CLS] token are fed to a standard Transformer network with a “pre-norm” layer normalization. The Transformer consists of a sequence of self-attention and feed-forward layers, paralleled with skip connections \cite{he2016deep}, which directly concatenate/merge features from different layers and enables gradients to flow better. The self-attention layers update the token representations by looking at other token representations with an attention mechanism. 

\paragraph{Attention Mechanism} Self-attention layers are the core component of Vision Transformers, which assign a pairwise attention value between every two tokens \cite{vaswani2017attention}. Each patch embedding builds its representation by “attending” to the other patches. By connecting other potentially distant patches of the image, the network builds a rich, high-level semantic understanding of the scene from the distributed patches in different space locations. Each attention head gathers relevant information from the transformed input vectors in different semantic spaces. Also, by visualizing the averaged attention maps among the heads of the last attention layer, we can see that the global attention map corresponds to coherent semantic regions of the raw input image at the top left plot of Appendix Figure \ref{fig:multi-head}.

\subsubsection{Supervised Multi-label Finetuning}
As the teacher network performs better than the student throughout the training, we extract the teacher network after the self-supervised pretraining is finished and add a linear classifier head for multi-label classification. Instead of using softmax activation-based Cross Entropy Loss, we chose the sigmoid activation-based Binary Cross Entropy With Logits Loss as our loss function for multi-label classification. We calculate the per-instance accuracy ($Acc = \frac{TP+TN}{FP+FN+TP+TN}$) by summing the number of true positives and true negatives divided by the number of dataset labels and averaging the final accuracy across the whole testing dataset.

\subsection{Explainer: Smooth Noise Norm Attention (SNNA)}

Mathematically, Attention computes output vector $\boldsymbol{y}^l\in\mathbb{R}^{(n+1)\times d}$ from a sequence of transformed vector $\boldsymbol{Z}^{l-1} = \{\boldsymbol{z^{l-1}_0}, ...,\boldsymbol{z^{l-1}_n}\} \subseteq \mathbb{R}^{d}$, where $n$ is the number of patches, $l \in\{1,...,L\}$, $L$ is the number of layers, and $d$ is the model embedding dimensions.

\begin{equation}\label{eq:1}
A^{l}_{i, j}\coloneqq \operatorname{softmax}\left(\frac{\boldsymbol{q}\left(\boldsymbol{z}^{l-1}_i\right) \boldsymbol{k}\left(\boldsymbol{z}^{l-1}_j\right)^{\top}}{\sqrt{d_{k}}}\right) \in \mathbb{R}
\end{equation}

$A_{i,j}^l$ is the attention weight token $\boldsymbol{z}^{l-1}_i$ assigned to token  $\boldsymbol{z}^{l-1}_j$.  $ A^l \in \mathbb{R}^{h \times (n+1) \times (n+1)}$, where $h$ is the number of heads. 

\begin{equation}\label{eq:2}
\begin{aligned}
& \boldsymbol{q}\left(\boldsymbol{z}^{l-1}_i\right):=\boldsymbol{z}^{l-1}_i \boldsymbol{W}^Q+\boldsymbol{b}^Q \quad\left(\boldsymbol{W}^Q \in \mathbb{R}^{d \times d_{k}}, \boldsymbol{b}^Q \in \mathbb{R}^{d_{k}}\right) \\
& \boldsymbol{k}\left(\boldsymbol{z}^{l-1}_j\right):=\boldsymbol{z}^{l-1}_j \boldsymbol{W}^K+\boldsymbol{b}^K \quad\left(\boldsymbol{W}^K \in \mathbb{R}^{d \times d_{k}}, \boldsymbol{b}^K \in \mathbb{R}^{d_{k}}\right) \\
& \boldsymbol{v}\left(\boldsymbol{z}^{l-1}_j\right):=\boldsymbol{z}^{l-1}_j \boldsymbol{W}^V+\boldsymbol{b}^V \quad\left(\boldsymbol{W}^V \in \mathbb{R}^{d \times d_{v}}, \boldsymbol{b}^V \in \mathbb{R}^{d_{v}}\right) \\
&
\end{aligned}
\end{equation}

$\boldsymbol{q(\cdot)},\boldsymbol{k(\cdot)}$ and $\boldsymbol{v(\cdot)}$ are query, key, and value transformations respectively. $\boldsymbol{W}^Q$,$\boldsymbol{W}^K$ and $\boldsymbol{W}^V$ are weight matrices used to project the input embeddings into query (Q), key (K) and value (V) vectors. 

\begin{equation}\label{eq:3}
\boldsymbol{y}^{l}_i=\left(\sum_{j=1}^n A^{l}_{i, j} \boldsymbol{v}\left(\boldsymbol{z}^{l-1}_j\right)\right) \boldsymbol{W}^O 
\end{equation}

Attention gathers value vector $\boldsymbol{v}\left(\boldsymbol{z}^{l-1}_j\right)$ based on attention weights $A_{i,j}^l$ and then projects them back to the embedding dimension $d$ with $\boldsymbol{W}^O \in \mathbb{R}^{d_{v} \times d}$ to get the multi-head attention output $\boldsymbol{y}^{l}_i$. As the matrix product is linear, Equation \ref{eq:3} can be rewritten as

\begin{equation}\label{eq:4}
\boldsymbol{y}^{l}_i=\sum_{j=0}^n A^{l}_{i, j}\hat{\boldsymbol{v}}_j^l 
\end{equation}

\begin{equation}\label{eq:5}
\hat{\boldsymbol{v}}^l:=(\boldsymbol{z}^{l-1}\boldsymbol{W}^V + \boldsymbol{b}^V)\boldsymbol{W}^O
\end{equation}

With this simple reformulation, we observe that the attention mechanism computes the weighted sum of the transformed value vectors from Equation \ref{eq:4} and \ref{eq:5}.

\begin{figure}[ht!]
\centering
\includegraphics[scale=0.52]{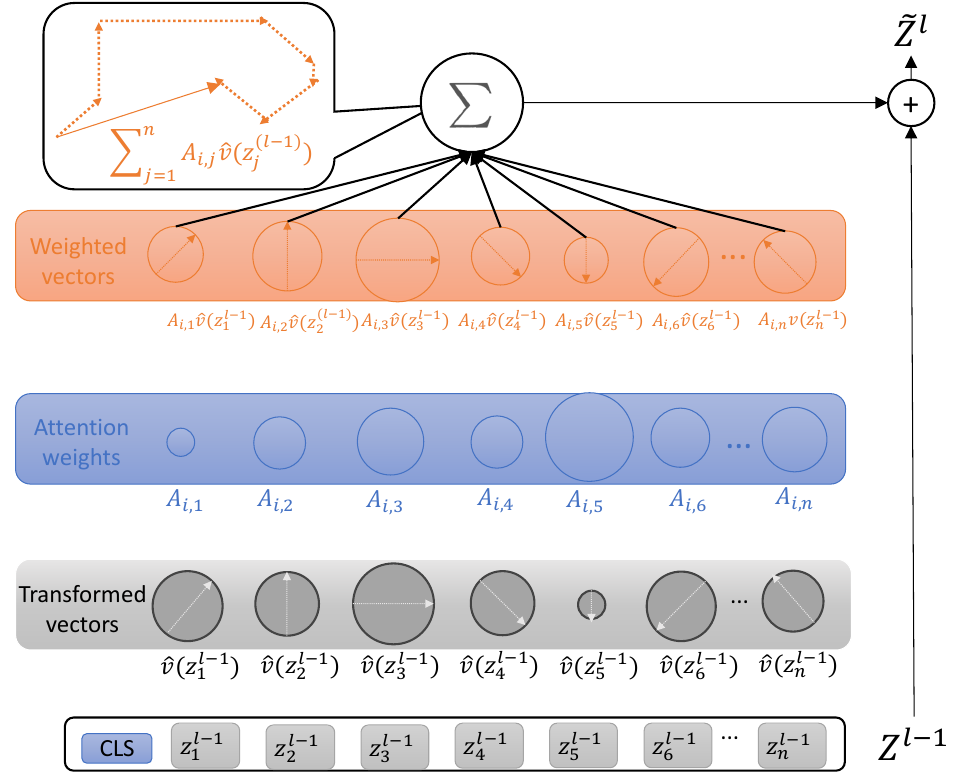}
\caption{Transformer Attention Mechanism: The size of the coloured circles depicts either the magnitude of the attention or the norm of the associated value vector; the larger the area of the circle, the bigger the value of the corresponding term. The arrows contained within the circles indicate the orientations of these vectors.}
\label{fig:attention}
\end{figure}

Previous works based on attention weight assume that the larger the attention weight of input vector $z_i^{l-1}$, the higher its contribution to the output $\Tilde{Z}^l$. However, this assumption disregards the magnitudes of the transformed value vector $\hat{\boldsymbol{v}}^{l-1}_i$. Neglecting the effect of the transformed value vector leads to the following problem as illustrated in Figure \ref{fig:attention}. 
For instance, the transformed value vector $\hat{v}(z_5^{l-1})$ is considerably smaller than the attention weight of $A_{i,5}$, resulting in the weighted vector $A_{i,5}\hat{v}(z_5^{l-1})$ contributing little to the output vector $\Tilde{Z}^l$. 
On the contrary, even the attention weight $A_{i,3}$ is not the largest. But, The transformed value vector $\hat{v}(z_3^{l-1})$ is large, leading to the weighted vector $A_{i,3}\hat{v}(z_3^{l-1})$ making a big contribution to the output vector $\Tilde{Z}^l$. Therefore, only considering the attention weight, in this case, may lead to the wrong interpretation of the input vector contribution to the output. 

\paragraph{Norm Weighted Attention} 
From Equation \ref{eq:4}, the space complexity of $A^{l}\hat{\boldsymbol{v}}^l$ ($ A^l \in \mathbb{R}^{h \times (n+1) \times (n+1)}$, $\boldsymbol{\hat{v}^l} \in \mathbb{R}^{(n+1) \times d}$)  is $O(h\times (n+1)^2 \times d)$. This operation is computationally expensive as the embedding dimension $d$ is usually large ($d=384$ in ViT-S). Herein, we propose using the norm weighted attention $A^{l}\|\boldsymbol{\hat{v}^l} \Vert_2$ ($\|\boldsymbol{\hat{v}^l} \Vert_2 \in \mathbb{R}^{n+1}$) to measure the magnitude of the input vector for the output, with the space complexity reduced to $O(h\times (n+1)^2)$.

\paragraph{Class Activation Map} \cite{zhou2016learning} To achieve class discriminative explanations, we define $\nabla A^{l} \coloneqq \frac{\partial f_c}{\partial A^{l}}$ as a mask and dot product it with $A^{l}\|\boldsymbol{\hat{v}^l} \Vert_2$, where $f_c$ is the model's output prediction on class $c$, and $\partial A^{l}$ is the raw attention of layer $l$. As we are more interested in positive relevance, so only the positive values of the "gradients-relevance multiplication" were considered when computing the relevance of weighted attention (denoted as "superscript $+$"). We treat the multiple heads equally, i.e., averaging multiple heads evenly. Considering the skip connections in the Transformer block, an identity matrix is added at the end of Equation \ref{eq:6} to account for residual connections \cite{abnar2020quantifying}.

\begin{equation}\label{eq:6}
\Bar{A}^{l} = \mathbb{E_h}((A^{l}\|\boldsymbol{\hat{v}^l} \Vert_2 \odot \nabla A^{l})^{+})+ \mathbb{1}
\end{equation}
Where $\Bar{A}^{l} \in \mathbb{R}^{(n+1) \times (n+1)}$, $\odot$ is the Hadamard product. $\mathbb E_h$ is the expectation across the head dimension.

\begin{figure}[ht]
\centering
\includegraphics[scale=0.45]{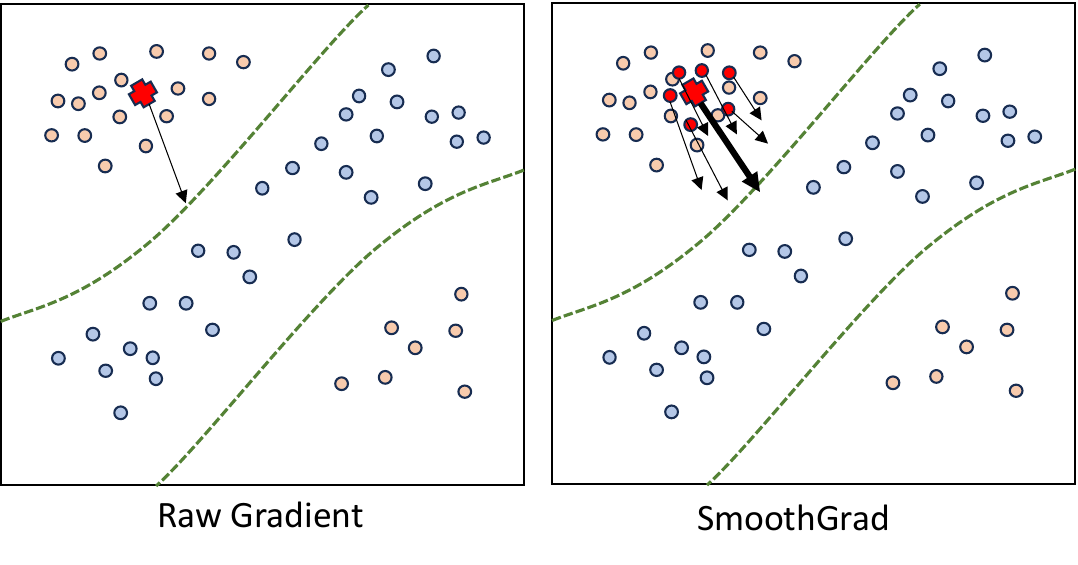}
\caption{A simple illustration of how SmoothGrad works compared to Raw Gradient. The plot shows a binary classification among light-blue and yellow-orange dots, where the \textbf{red cross} is the test sample, and the green dashed line represents the learned decision boundary. Compared with Raw Gradient (slim arrow), SmoothGrad samples points (red dots) around the test point and calculates the averaged Gradient (bold arrow). }
\label{fig:SmoothGrad}
\end{figure}
 
\paragraph{Smooth Noise} To avoid highlighting irrelevant pixels and smoothen the attribution maps, we draw inspiration from SmoothGrad \cite{smilkov2017smoothgrad}: "removing noise by adding noise", which uses a random sampling strategy around the input by averaging the obtained attributions to produce visually sharper attribution map, as illustrated in Figure \ref{fig:SmoothGrad}. The smoothing process effectively retains the relevant parts and reduces the gradient self-induced noise \cite{goh2021understanding}, which refers to the instability introduced during the backpropagation, including numerical instability and vanishing or exploding gradients. When these sources of noise or instability affect the gradients computed during backpropagation, they can lead to inaccurate or unreliable explanations. We smooth the raw gradients over the input space, which computes an attribution map by averaging multiple attribution maps with $m$ permutation inputs.

\begin{equation}\label{eq:7}
SG(x_\epsilon)= \frac{1}{m}\sum_{1}^m\frac{\partial{f_c(x_\epsilon)}}{\partial{A^L}}
\end{equation}

Where $x_\epsilon=x +\mathcal{N}(0, \sigma^2)$, $x$ is the input image, and $A^L$ is the last layer raw attention. Gaussian noise $\mathcal{N}$ is used to smoothen the input space and construct visually sharper attribution maps. It is briefly discussed in \cite{smilkov2017smoothgrad} that standard deviation $\sigma$ needs to be carefully selected to get the best result. If too small, the attribution maps are still noisy; if too large, the maps become irrelevant. We empirically found that $m=5,\sigma=0.15(\max(x)-\min(x))$ is sufficient in this work. Then, we multiply the per layer norm weighted attention and element-wise product of the SmoothGrad to get the relevance matrix $\mathbf{R}(x)$. By weighting the attention, noise in the model amplified/caused by applying the multi-layer transition process is reduced or eliminated.

\begin{equation}\label{eq:8}
\mathbf{R}(x) = (\Bar{A}^{1}\cdot\Bar{A}^{2}\dots\cdot\Bar{A}^{L}) \odot SG(x_\epsilon)
\end{equation}

In ViT, as shown in Figure \ref{fig:attention} the [CLS] token serves as a class token, aggregating information from the entire input sequence that encapsulates global context information of the image. To retrieve per-token relevance for classification tasks, we take the row corresponding to [CLS] token, namely $R[0]$, then skip the first [CLS] token to get the final attribution vectors $ R[0, 1:]$ corresponding to the input patch sequences. We generate the visualization map by reshaping the attribution vectors to a matrix, then up-sampling it back to the original input image size by bilinear interpolation with the scale factor of patch size $p$, as is illustrated in the Appendix Algorithm \ref{algom:1}.

\section{Experiments}

\subsection{Dataset}
The Berkeley Deep Drive dataset \footnote{\url{https://doc.bdd100k.com/download.html}}(BDD-100K)\cite{yu2020bdd100k} contains 100,000 images with 2D bounding boxes, lane markings, and full-frame instance segmentation. As an extension of BDD-100K, BDD-OIA \footnote{\url{https://twizwei.github.io/bddoia_project/}} \cite{xu2020explainable} collected images of complicated scenes (\#pedestrians > 5 or \#vehicles > 5) from the original BDD-100K dataset and then annotated them with four action categories: move forward, slow down, turn left and turn right. The total 22k annotated images are split into subsets of 16k for the training, 2k for the validation and 4k for the test set. To our knowledge, BDD-OIA is the only publicly available annotated driving action dataset with binary labelling: potentially executable actions are marked as 1, while the rest are marked as 0, e.g., [1,1,0,0], which means you can move forward or slow down. Samples of the images and their annotation can be seen in Appendix Figure \ref{fig:BDD-OIA}. Both the aforementioned datasets are used in this work. The size of the RGB image in both datasets equals $720\times1280$. 

The model training details can be found in Appendix \ref{Imp_detail}.

\subsection{Baseline explainable methods}
We chose the four currently most representative attention-based explainable methods plus one self-proposed method, i.e., AttIG, and compared SNNA with them through qualitative and quantitative experiments.
\textbf{RawAtt}\cite{jain2019attention} utilize the raw attention weights from the last layer $A^{L}$, but can only get class-agnostic attribution.
\textbf{AttGrad}\cite{DBLP:conf/acl/ChrysostomouA20} regard the partial derivative of the output to the input as a measure of the network's sensitivity for each input dimension and element-wise product of the attention matrix with Attention Gradient as the final relevancy map $ A^L \odot \nabla {A^L}$.
\textbf{AttIN}\cite{kobayashi2020attention} multiply the attention matrix with the transformed value vectors, then take the norm of the as the final attribution: $\|A^L\hat{v}^L\Vert$. 
\textbf{GenericAtt}\cite{chefer2021generic} combine Attention Gradient and Attention Rollout, which consider only the positive values of the gradients-relevance multiplication and average the result across the “heads” dimension, $A^{*} = \prod_{i=1}^L(\mathbb{E_h}(A^i \odot \nabla A^i)^{+} + \mathbb{1})$.
\textbf{AttIG} is a modification of the GenericAtt, which masks $A^*$ by $IG(x) = (x-x') \times \int^1_0 \frac{\partial f(x' + \alpha(x-x'))}{\partial x} d\alpha$, where $x$ is the original input and $x'$ is the baseline point.


\subsection{Qualitative evaluation}

The qualitative evaluation is based on the inspection of the produced attribution maps. However, this introduces a strong bias in the evaluation. Humans judge methods more favourably that produce explanations closer to their expectations at the cost of penalizing methods that might more closely reflect the network behaviour. There are limitations to the qualitative evaluation of attribution maps due to biases in human intuition towards simplicity. Therefore, apart from qualitative comparison, we use quantitative metrics defined in Section \ref{metrics} as a complementary for more trustworthy evaluation.

\subsection{Quantitative evaluation}\label{metrics}
A common method in model interpretability involves systematically removing or altering input features and observing the impact on the model's output. The significance of input features for the final decision can be measured by the change in model performance without them. Removing features with high attribution scores should decrease the performance, while discarding features with low attribution scores has less effect on the model's performance \cite{wang2024visual}.

\paragraph{Faithfulness} quantifies the fidelity of an explanation technique by measuring if the identified tokens impact the output. We regard the attribution value as a relevance score; the larger the value, the stronger the correlation, and perform positive perturbation tests: tokens are removed from the highest relevance to the lowest, to evaluate the explanation faithfulness by Area-under-the-perturbation-curve (AUPC) \cite{gupta2022new} and Log-odds scores \cite{qiang2022attcat} metrics. We gradually "remove" the input features of a given input and measure the accuracy of the network. In positive perturbation, We expect a steep decrease in performance, indicating that the removed tokens are important to the classification score.

\paragraph{AUPC} measures the total area under the accuracy perturbation curve. Here, we calculate the average perturbation accuracy overall test examples under masking of the top $k\%$ pixels/token/attention, then approximate the AUPC by summing over the candidates $k$ set.

\begin{equation}\label{eq:11}
    AUPC=\sum_{k}\frac{1}{N}\sum_{i=1}^{N}Acc(f(\Tilde{x}_i^k))
\end{equation}

\paragraph{LogOdd} calculates the average difference of logarithmic probabilities after and before masking the top $k\%$ pixels/token/attention over all test examples, then sum over the candidates $k$ set.

\begin{equation}\label{eq:12}
    LogOdd=\sum_{k}\frac{1}{N}\sum_{i=1}^{N}\log \frac{Acc(f(\Tilde{x}_i^k))}{Acc(f(x_i))}
\end{equation}

$N$ is the number of test samples, $x_i$ is the test sample, and $\Tilde{x}_i^k$ is the masked sample after masking the top $k\%$ pixels/token/attention. $Acc$ is the multi-label classification accuracy. $\sum_{k}$ means sum over candidates $k$ set. Directly removing specified tokens from the original input results in a shorter length of the new input patches sequence. Which may cause the out-of-distribution (OOD) problem. Regarding OOD problems in the ablation examination of \citet{hooker2019benchmark}, we utilize the below three replacement methods suggested by \citet{hase2021out}. Namely, Pixel mask, Token mask and Attention mask.

\begin{figure*}[ht!]
    \centering
    \includegraphics[width=\textwidth]{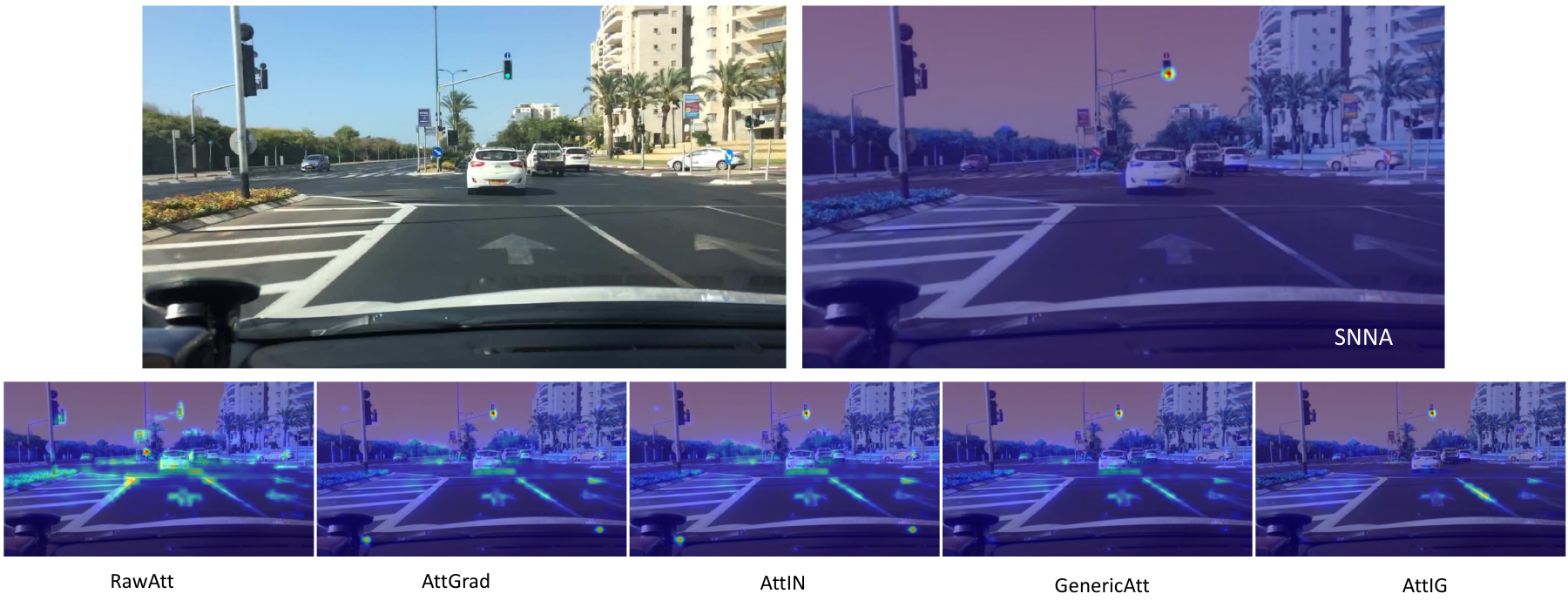}
    \vspace{-1em}
    \caption{Explainable results: our method produces a more reliable and clear visualization Heatmap, with the predicted action of "move forward". }
    \label{fig:compare}
\end{figure*}

\subsection{Replacement Methods}

\paragraph{Pixel Mask} For ViT, the input image will undergo normalization as part of pre-processing. This means that ViT does not actually "see" the raw pixel values; instead, it sees normalized values, where the mean pixel value of the image dataset might not be 0. Therefore, setting the pixel value to 0 (black) could introduce a significant statistical outlier for the model rather than a neutral or "uninformative" input. It can introduce biases and disrupt the context or texture continuity of the surrounding area, potentially leading the model to focus on the edges of the masked area rather than ignoring it. So, we replace the corresponding patch pixel with the per channel-wise averaged value of the input image.

\paragraph{Token Mask} Setting the token value of a specific patch to the zero vector means essentially removing the information associated with that patch from the input representation \cite{li2016understanding}. This forces the model to redistribute its attention to the remaining non-zero tokens. Using a zero vector to replace a patch's embedding is straightforward and does not disrupt the input's expected shape or the subsequent processing pipeline. As a result, we can maintain the original dimensional structure of the input and ensure that the model architecture does not need any modification to accommodate the masked input. Hence, we replace the image patch tokens with the zero vector.

\paragraph{Attention Mask} Sets the attention weights for the masked tokens to 0, so the masked parts will not be forwarded in the network anymore \cite{serrano2019attention}. This approach is grounded by the attention mechanism, where attention weights determine the importance of each token when computing the output of the attention layers. By setting the attention weights of certain tokens to zero, we effectively remove their influence on the subsequent layers, allowing us to assess how critical they are for the model's decision.

\begin{figure*}[ht!]
    \centering
    \includegraphics[width=\textwidth]{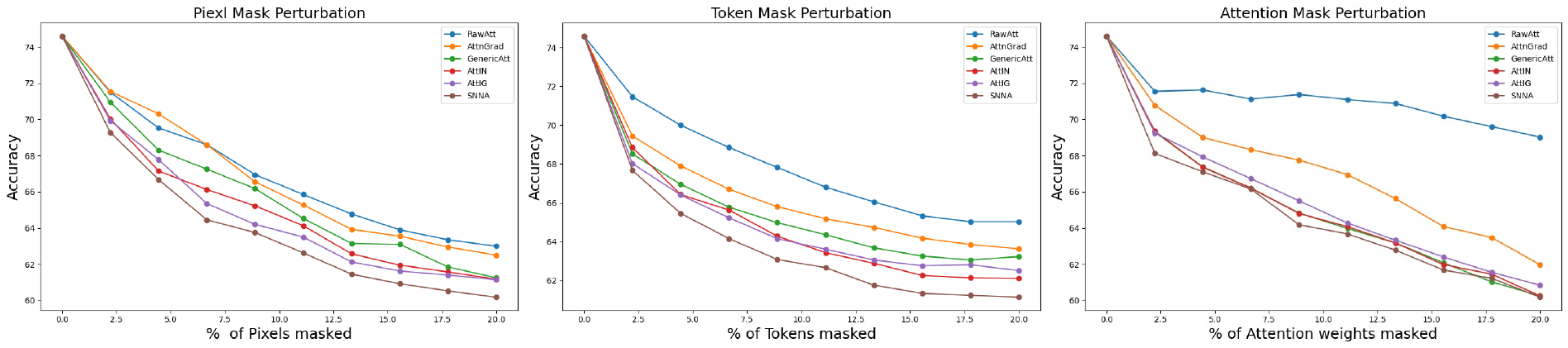}
     \vspace{-1em}
    \caption{Perturbation experiments result of Pixel mask, Token mask and Attention mask. The smaller the area under the curve, the more effective the corresponding method is. The x-axis demonstrates masking the $Top-k\%$ of Pixel/Token/Attention in decreasing order. }
    \label{fig:perturbation}
    \vspace{1em}
\end{figure*}

\begin{table*}[ht!]
\caption{The results in this Table were calculated by AUPC \ref{eq:11} and LogOdd \ref{eq:12} metrics, $\downarrow$ means the lower the better.}
\vspace{1em}
\resizebox{2 \columnwidth}{!}
{
\centering
\begin{tabular}{|c|c|c|c|c|c|c|c|c|c|c|c|c|}
\hline
\multirow{2}{*}{Method} & \multicolumn{2}{c|}{RawAtt} & \multicolumn{2}{c|}{AttGrad} & \multicolumn{2}{c|}{AttIN} & \multicolumn{2}{c|}{GenericAtt} & \multicolumn{2}{c|}{AttIG} & \multicolumn{2}{c|}{SNNA}  \\
\cline{2-13}
& AUPC$\downarrow$ & LogOdd$\downarrow$& AUPC$\downarrow$ & LogOdd$\downarrow$ & AUPC$\downarrow$& LogOdd$\downarrow$ & AUPC$\downarrow$ & LogOdd$\downarrow$ & AUPC$\downarrow$  & LogOdd$\downarrow$ & AUPC$\downarrow$ & LogOdd$\downarrow$ \\
\hline
Pixel mask & 672.95 & -1.06 & 669.8 & -1.09 & 661.15 & -1.23 & 654.52 & -1.33& 651.65 & -1.37 & \textbf{644.45} & \textbf{-1.49} \\
\hline
Token mask & 680.97 & -0.92& 666.0& -1.15 & 658.38 & -1.26 & 652.55 & -1.35& 653.1 & -1.34 & \textbf{643.02} & \textbf{-1.5} \\
\hline
Attention mask & 711.05 & -0.48 & 672.55 & -1.05 & 652.8 & -1.35 &  653.22 & -1.35 & 656.4 & -1.3 & \textbf{649.67} & \textbf{-1.4} \\
\hline
\end{tabular}
}
\label{table:metrics}
\end{table*}

\section{Results}

\subsection{Qualitative}
Figure \ref{fig:compare} presents a visual comparison between SNNA and various other baselines. Where RawAtt create a very scattered saliency map, many unrelated areas are marked as significant, like the indicator lights on the far left of the picture, as well as the flower beds and the traffic sign in the distance. AttGrad and AttIN have similar results, but strangely, two small areas at the bottom of the picture are highlighted to the left and right front of the car. Overall, the results of GenericAtt and AttIG are the best among baselines, but they still have some puzzling noise. From Figure \ref{fig:compare}, we can see that our method provides clearer and more intuitive visualization with respect to the baselines, which produce more or less noise attribution. More convincing visual results can be found in Appendix Figure \ref{fig:appendix}.

\subsection{Quantitative}
Due to computational costs, all the experiment results are averaged on 1000 samples randomly chosen from the BDD-OIA testing dataset. As the saliency pixel in driving scenarios normally accounts for a relatively small area, we focus our ablation experiment on the top $20\%$ pixels/token/attention.  It can be seen from Table \ref{table:metrics} that SNNA achieves the lowest AUPC and LogOdd scores in most settings, demonstrating that SNNA can pick out the most impactful pixels for model prediction. RawAtt has the worst performance in both metrics among all the compared methods. This demonstrates the inadequacy of purely relying on attention alone, regardless of the magnitude of the input features. The effect of Pixel Mask and Token Mask is roughly comparable. Interestingly, the RawAtt line is less sensitive to the attention mask from Figure \ref{fig:perturbation}, implying that RawAtt may not be reliable. Overall, the quantitative evaluations demonstrate that SNNA outperforms most baseline methods.

\section{Conclusion}
This study addresses significant challenges in generating clear and confident explanations for Transformer-based models by introducing a novel approach termed Smooth Noise Norm Attention (SNNA). It utilizes attention weights, value vectors, and gradients to calculate attribution scores and quantify input features' influence on model outputs. These attribution scores provide the significance of the input features for the model's decision. Our experimental results reveal that SNNA can produce clear attribution maps that are more relevance-sensitive and with much less noise than baseline methods. We also demonstrate that SNNA outperforms baseline methods in ranking the importance of the input pixels regarding the quantitative metrics. Although our current SNNA method is specifically implemented on ViT for image classification tasks, it can be naturally applied to any Transformer-based architectures in future research. As there is no ground truth annotation for the "noisy" pixel, the investigation regarding the faithfulness of the salient pixels to the model's decision remains to be investigated in future research. Due to the lack of credible evaluation metrics, how to fairly and effectively assess the effectiveness of XAI methods remains to be explored.


\begin{ack}
The work was partially funded by the following grants: The Horizon MSCA projects PERSEO and TRAIL, UKRI TAS node on Trust, ERC Advanced eTALK (funded by UKRI), and US AFOSR project CASPER++.
We would like to acknowledge the assistance given by Research IT and the use of the Computational Shared Facility at The University of Manchester. We thank Yuping Wu and Jacopo de Berardinis for their constructive suggestions.
\end{ack}



\bibliography{mybibfile}

\onecolumn
\appendix

\section{Appendix}

\subsection{Implementation details} \label{Imp_detail}

We train the classification model on the previously introduced two datasets. We chose ViT-S/8 with L = 12 Transformer encoder layers for this work. We first initialize our model from the pretrained checkpoint (ViT-S/8) on ImageNet and train it on the unlabelled BDD-100k training dataset with a batch size of 16 for 200 epochs on two Nvidia A100 (80GB) GPUs. Then, we freeze the backbone network and fine-tune the classification head with the labelled data for 100 epochs (Suggested by DINO\footnote{\url{https://github.com/facebookresearch/dino}}). During pretraining, we follow the suggestion of DINO to randomly crop and then resize the original image to two $224 \times 224$ bigger crops and eight $96 \times 96$ small crops for the teacher and student network separately. Considering the computational resource limits, we resized the input image to $360\times640$ during supervised fine-tuning. Our final trained driving-action prediction model gets $74.6\%$ multi-label classification accuracy on the BDD-OIA testing dataset, which is higher than the baseline \cite{xu2020explainable} accuracy of $73.4\%$.

\begin{algorithm}
\caption{Generate Visualization}
\begin{algorithmic}[1]
\Procedure{GenerateVis}{image, patch\_size, attribution}
    \State $w, h \gets \text{dimensions of } image$
    \State $w^* \gets w \mod patch\_size$, $h^* \gets h \mod patch\_size$
    \State $mask \gets reshape(attribution,(w^*,h^*))$
    \State $mask \gets interpolate(mask, patch\_size)$ 
    \State $mask \gets \text{normalize}(mask)$ \Comment{Min-Max Normalize}
    \State \Comment{chage grayscale mask to colormap}
    \State $heatmap \gets applyColorMap(255*mask, colortype)$ 
    \State $vis \gets heatmap + image$ \Comment{Overlay heatmap on image}
    \State \textbf{return} $vis$
\EndProcedure
\end{algorithmic}
\label{algom:1}
\end{algorithm}

\begin{figure}[hbt!]
\centering
\includegraphics[scale=0.54]{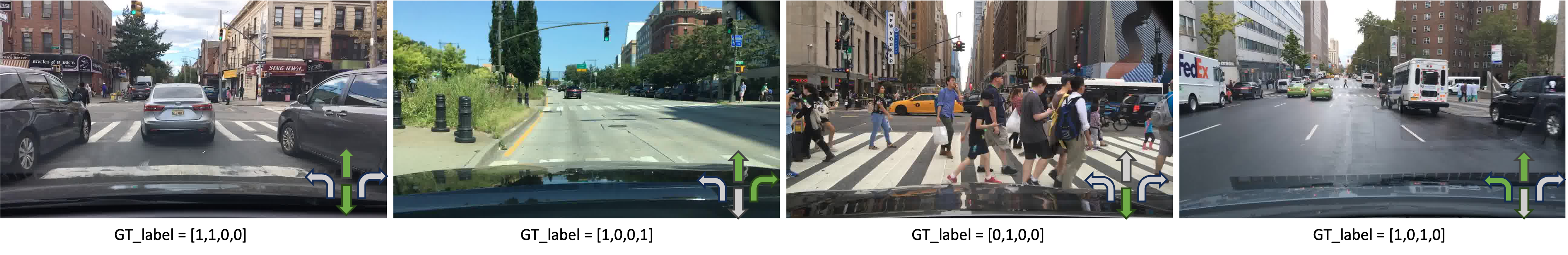}
\caption{Image samples from BDD-OIA dataset. The green arrows show the ground truth label for possible driving actions, and the ground truth binary vector is below the image.}
\label{fig:BDD-OIA}
\end{figure}

\begin{figure}[hbt!]
    \centering
    \includegraphics[scale=0.54]{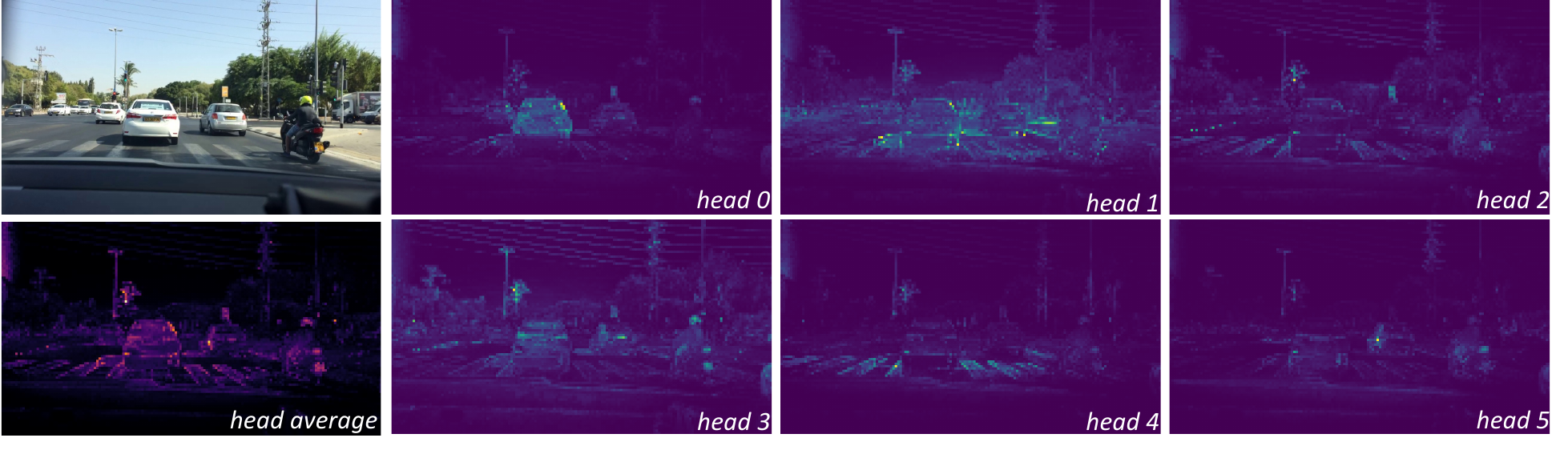}
    \caption{Attention maps from multiple heads. We consider the heads from the last layer of a ViT-S/8 trained with DINO and display the self-attention for [CLS] token query. It can be seen different heads focus on different locations that represent different objects or parts.}
    \label{fig:multi-head}
\end{figure}

\begin{figure}[ht!]
    \centering
    \includegraphics[scale=0.85]{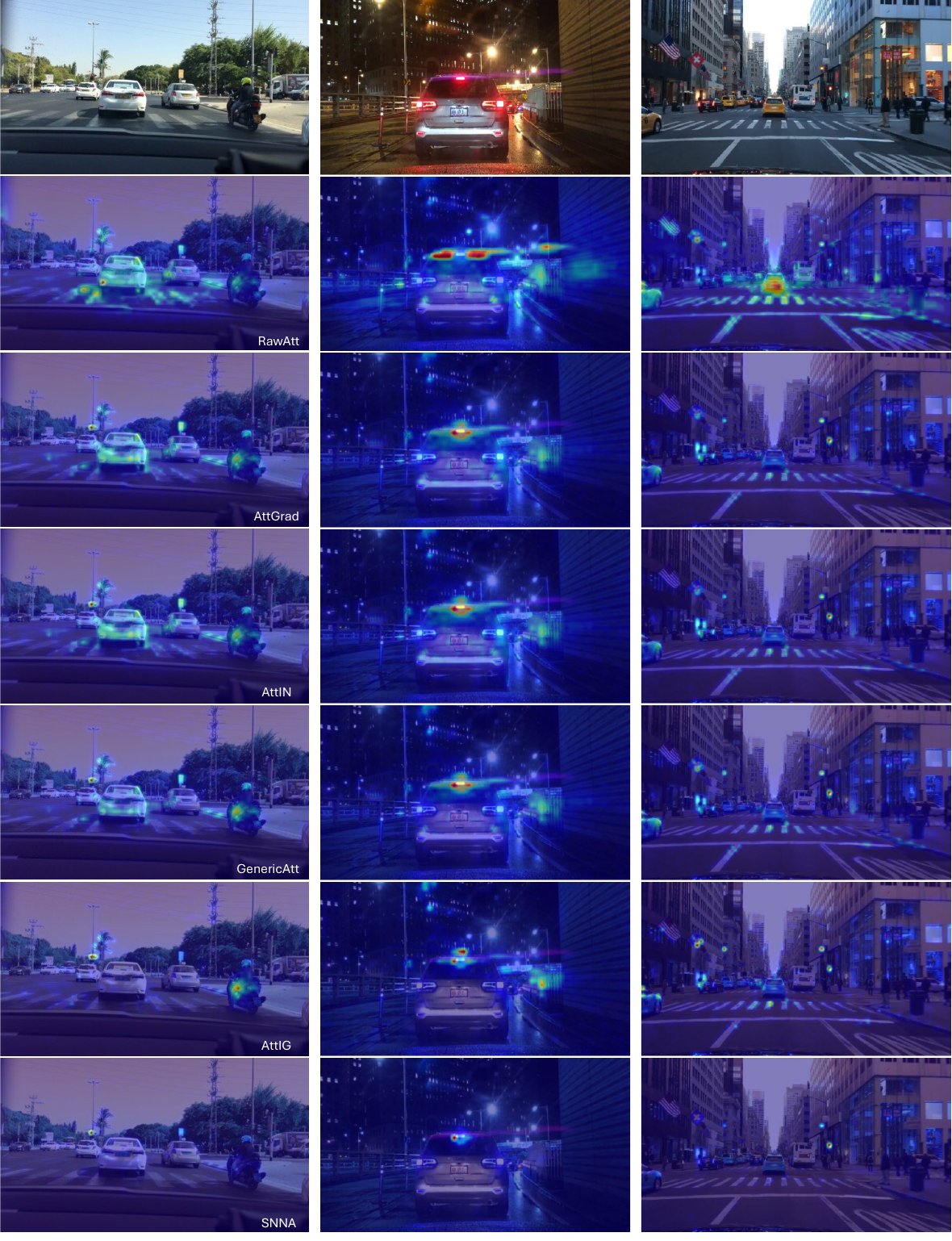}
    \caption{More sample results, images from top to bottom corresponding to the input image, RawAtt, AttGrad, AttIN, GenericAtt, AttIG and SNNA. The predicted actions of those images are 'move forward', 'slow down' and 'slow down' from left to right. }
    \label{fig:appendix}
\end{figure}

\end{document}